\documentclass[conference]{IEEEtran}
\IEEEoverridecommandlockouts
\usepackage{amsmath,amsfonts}
\usepackage{array}
\usepackage{textcomp}
\usepackage{stfloats}
\usepackage{url}
\usepackage{verbatim}
\usepackage{graphicx}
\usepackage{amssymb}
\usepackage{hyperref}
\usepackage{graphicx}
\usepackage{amsmath}
\usepackage{longtable}
\usepackage{algorithm} 
\usepackage{algpseudocode} 
\usepackage{mathrsfs}
\usepackage{subcaption}
\usepackage{mathtools}
\usepackage{pifont}
\usepackage{color}
\usepackage{lineno}
\usepackage{makecell} 
\usepackage{pdflscape}
\usepackage{adjustbox}
\usepackage[utf8]{inputenc}
\usepackage{tabularx}
\usepackage{blindtext}
\usepackage{longtable}
\usepackage{lscape}
\usepackage{setspace}
\usepackage{graphicx}
\usepackage{multirow}

\usepackage{notoccite} 
\usepackage{lscape} 
\usepackage{mwe}
\usepackage{booktabs}
\usepackage{amsthm}

\theoremstyle{definition}

\theoremstyle{definition}

\newcommand{\RNum}[1]{\lowercase\expandafter{\romannumeral #1\relax}}
\newcommand{\RNumU}[1]{\uppercase\expandafter{\romannumeral #1\relax}}
\usepackage[numbers]{natbib}
\def\BibTeX{{\rm B\kern-.05em{\sc i\kern-.025em b}\kern-.08em
    T\kern-.1667em\lower.7ex\hbox{E}\kern-.125emX}}
\usepackage{balance}
\usepackage{hyperref}
\begin{document}
\title{Robust Broad Learning System with Wave Loss for Classification under Data Uncertainty}
\author{
\IEEEauthorblockN{Mushir Akhtar}
\IEEEauthorblockA{
\textit{Department of Mathematics} \\
\textit{Indian Institute of Technology Indore}\\
phd2101241004@iiti.ac.in}
\and
\IEEEauthorblockN{Abhishek Varshney}
\IEEEauthorblockA{
\textit{Department of Mathematics} \\
\textit{Indian Institute of Technology Indore}\\
phd2401141003@iiti.ac.in}
\and
\IEEEauthorblockN{A. Quadir}
\IEEEauthorblockA{
\textit{Department of Mathematics} \\
\textit{Indian Institute of Technology Indore}\\
mscphd2207141002@iiti.ac.in}
\and
\IEEEauthorblockN{A. Rahaman}
\IEEEauthorblockA{
\textit{Department of Mathematics} \\
\textit{Indian Institute of Technology Indore}\\
phd2401141001@iiti.ac.in}
\and
\IEEEauthorblockN{M. Tanveer}
\IEEEauthorblockA{
\textit{Department of Mathematics} \\
\textit{Indian Institute of Technology Indore}\\
mtanveer@iiti.ac.in}
\and
\IEEEauthorblockN{Mohd. Arshad}
\IEEEauthorblockA{
\textit{Department of Mathematics} \\
\textit{Indian Institute of Technology Indore}\\
arshad@iiti.ac.in}
}
\maketitle
\begin{abstract}
Broad Learning System (BLS) offers an efficient alternative to deep architectures by enabling fast learning through randomized feature mapping and closed-form solutions. However, its reliance on squared error loss makes it highly sensitive to noise, outliers, and corrupted labels, limiting its reliability in real-world scenarios. To address this limitation, we propose Wave-BLS, a robust broad learning framework that integrates the wave loss function, which is asymmetric, bounded, and smooth, enabling controlled penalization of large errors. The proposed formulation replaces the standard least-squares objective with a wave-loss-based optimization problem, solved efficiently using a Nesterov accelerated gradient (NAG)-based scheme without requiring matrix inversion, thereby improving scalability. Extensive experiments on 30 UCI benchmark datasets demonstrate that Wave-BLS consistently outperforms classical BLS and several robust variants. Statistical validation using Friedman and Nemenyi post-hoc tests confirms the significance of the observed improvements. Furthermore, robustness evaluations under controlled noise and outlier injection reveal that Wave-BLS exhibits substantially slower performance degradation compared to BLS, even in challenging contamination settings. These results establish Wave-BLS as a stable and robust alternative to existing broad learning models for learning under data uncertainty.
\end{abstract}
\begin{IEEEkeywords}
Randomized Neural Network, Broad Learning System, Wave Loss Function, Robust Classification, Noise and Outlier Robustness
\end{IEEEkeywords}

\section{Introduction}


\IEEEPARstart{R}{andomized} neural networks (RaNNs) have emerged as efficient alternatives to deep neural networks, offering reduced computational complexity, fewer parameters, and less sensitivity to hyperparameter tuning while maintaining competitive performance \cite{zhang2016survey}.
By fixing randomly generated hidden-layer parameters and learning only the output weights in closed form, RaNNs enable fast training with reduced computational overhead. Among them, the random vector functional link (RVFL) network \cite{malik2023random} has gained prominence due to its direct input-output connections, which allow simultaneous modeling of linear and nonlinear relationships and yield strong generalization performance \cite{ AKHTAR2026112711}. 

Despite these advantages, the shallow architecture of RVFL limits its representational capacity for complex data distributions \cite{chen2017broad1}. To overcome this limitation, the Broad Learning System (BLS) was introduced as a flat yet expressive architecture that expands network width instead of depth \cite{chen2017broad2}. BLS constructs multiple groups of randomized feature and enhancement nodes whose outputs are concatenated to form a rich representation, while retaining fast training through a closed-form least-squares solution. This design avoids backpropagation, supports incremental learning, and performs well in limited-data regimes.

Nevertheless, BLS suffers from two fundamental limitations: 
(i) its dependence on the squared-error loss makes it highly sensitive to noise and outliers, and 
(ii) the computation of output weights relies on matrix inversion, resulting in cubic-time complexity that limits scalability.

A range of BLS variants have been proposed to improve robustness and mitigate sensitivity to noise. For example, \citet{jin2018regularized} introduced a regularized BLS (R-BLS) by incorporating $\ell_1$, $\ell_2$, and elastic-net penalties into the least-squares formulation, which helps control model complexity but remains sensitive to severe noise and outliers due to its reliance on the squared-error loss. To address uncertainty, \citet{feng2018fuzzy} proposed a neuro-fuzzy BLS (NF-BLS) that integrates fuzzy IF-THEN rules via k-means clustering, improving interpretability and robustness at the cost of additional computational overhead. More recently, the SC-KSBLS framework \cite{10902561} replaces the squared-error loss with a kernel risk-sensitive mean $p$-power loss and incorporates adaptive label correction. While effective under label noise, its collaborative structure and kernel-based optimization substantially increase computational cost. Additional extensions have addressed specific challenges such as imbalanced noisy data via graph-fuzzy embeddings \cite{10906533}, privacy-preserving multiparty learning \cite{10025360}, and fractional-order BLS formulations \cite{9784155}. Recently, copula-aligned weight initialization (CAWI) has been proposed to incorporate dependency-aware copula modeling into weight initialization, aligning randomized weights with the intrinsic structure of the input data and thereby improving representation quality and learning performance \cite{akhtar2026cawi}.

Overall, existing BLS variants enhance robustness through regularization, fuzzy inference, kernelization, or collaborative correction mechanisms, often at the expense of increased architectural complexity or computational overhead. These limitations motivate the need for a robust yet computationally efficient BLS framework that improves noise tolerance while preserving the simplicity of the original architecture.

The choice of loss function plays a central role in robustness against noise and outliers \cite{tian2022recent}, yet most BLS-based methods continue to rely on the conventional least-squares loss. While robustness can be improved through auxiliary structures such as clustering or fuzzy membership evaluation, these additions inevitably increase computational burden. In contrast, directly adopting a robust loss function within the BLS optimization framework offers a principled and efficient alternative. Recent advances demonstrate the effectiveness of this strategy across related learning settings. The RoBoSS loss combines boundedness, sparsity, and smoothness for robust supervised learning \cite{akhtar2025roboss}, while the HawkEye loss additionally incorporates an insensitive zone for noise-tolerant regression \cite{akhtar2025hawkeye}. Integrating HawkEye with an RVFL network has also produced a robust classifier for noisy and outlier-prone data \cite{akhtar2024advancingrvflnetworksrobust}. In the multiview setting, the wave loss has been used to improve robustness while jointly exploiting consensus and complementary information across views \cite{quadir2025multiviewsynergy}.

Motivated by this perspective, we consider the wave loss function \cite{akhtar2024advancing}, which has demonstrated strong robustness properties due to its boundedness, smoothness, and controllable asymmetry. We investigate whether the wave loss can be effectively integrated into the BLS framework to address its sensitivity to noise and outliers. Building on this idea, we propose Wave-BLS, i.e., a broad learning system with wave loss function, which replaces the traditional least-squares loss while retaining the flat architecture of BLS. In addition, we reformulate the parameter optimization process to eliminate the matrix inversion step required in conventional BLS, thereby significantly enhancing scalability. The resulting optimization problem is efficiently solved using a Nesterov accelerated gradient (NAG)-based algorithm, which ensures stable updates and fast convergence. In the following sections, we demonstrate that the proposed Wave-BLS model achieves improved generalization performance compared with existing baseline models. 

The key contributions of the work are as follows: 
\begin{enumerate}
    \item We propose Wave-BLS, a robust broad learning system that integrates the wave loss function to effectively mitigate the impact of noise and outliers.
    \item  We formulate the optimization framework in Wave-BLS to eliminate the matrix inversion step required in traditional BLS, significantly enhancing its scalability.
 \item We employ a NAG-based optimization algorithm to solve the proposed Wave-BLS model, ensuring efficient parameter updates and faster convergence.
\item We extensively evaluate the proposed Wave-BLS on 30 UCI benchmark datasets using accuracy and rank metrics, and further establish its statistical significance through Friedman and Nemenyi post-hoc tests.
\item We further assess the robustness of Wave-BLS under controlled data corruption by injecting varying levels of noise and outliers. The experimental results confirm that Wave-BLS maintains superior performance compared to the standard BLS.
\end{enumerate}

The remainder of this paper is organized as follows. Section~\ref{Related-Work} reviews the architecture and formulation of BLS. Section~\ref{Proposed-work} presents the proposed Wave-BLS model. Experimental evaluations are reported in Section~\ref{Experiment-section}. Finally, Section~\ref{Conclusions-section} concludes the paper.

\section{Related Work} \label{Related-Work}
This section introduces the notations and presents an overview of the architecture and formulation of BLS.

\subsection{Notations}
Let \( X \in \mathbb{R}^{m \times d} \) represent the input data matrix, where \( m \) is the number of samples and \( d \) is the number of features. The target matrix is denoted by \( Y \in \mathbb{R}^{m \times n_{\text{out}}} \), where \( n_{\text{out}} \) is the number of classes. The outputs from the feature layer are represented by \( Z \in \mathbb{R}^{m \times p q} \), where \( p \) and \( q \) denote the number of nodes per window and the number of feature node windows, respectively. Similarly, the outputs from the enhancement layer are represented by \( H \in \mathbb{R}^{m \times r s} \), where \( r \) and \( s \) denote the number of nodes per window and the number of enhancement node windows, respectively. The combined representation matrix, obtained by concatenating the feature and enhancement layer outputs, is denoted as \( A \in \mathbb{R}^{m \times (pq + rs)} \). Finally, the unknown weight matrix is denoted by \( W \in \mathbb{R}^{(pq + rs) \times n_{\text{out}}} \).

\subsection{Broad Learning System (BLS): Architecture and Formulation \cite{chen2017broad2}}

\begin{figure}[t]
    \centering
    \includegraphics[width=0.90\linewidth]{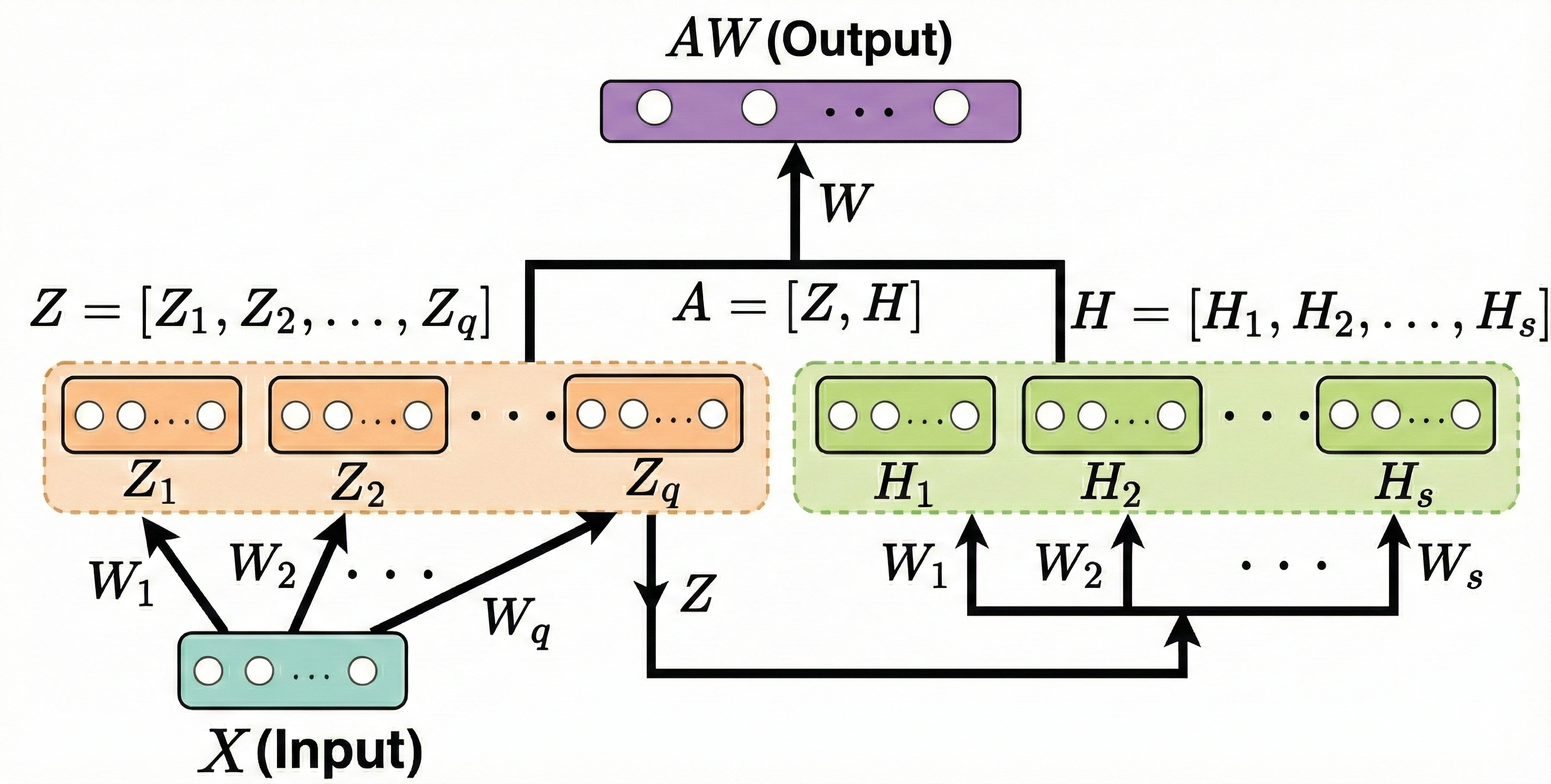}
    \caption{Architecture of the standard BLS.} 
    \label{fig:BLS_architecture}
\end{figure}

BLS is a flat network architecture composed of three main components: a feature layer, an enhancement layer, and an output layer. Unlike deep architectures, BLS expands the network width by incrementally adding feature and enhancement nodes, enabling efficient learning via closed-form solutions. Fig.~\ref{fig:BLS_architecture} illustrates the overall architecture of the BLS framework.

\noindent
\textbf{Feature Layer:}
Given an input data matrix \(X \in \mathbb{R}^{m \times d}\), the feature layer generates \(q\) groups (windows) of feature nodes, each containing \(p\) nodes. The output of the \(i^{th}\) feature window is computed as
\begin{align}
Z_{f_i} = \phi(X W_i + B_i),
\end{align}
where \(W_i \in \mathbb{R}^{d \times p}\) and \(B_i \in \mathbb{R}^{m \times p}\) are randomly initialized weights and biases, respectively, and \(\phi(\cdot)\) denotes a nonlinear activation function. Concatenating all feature windows yields
\begin{align}
Z = [Z_{f_1}, Z_{f_2}, \ldots, Z_{f_q}] \in \mathbb{R}^{m \times pq}.
\end{align}

\noindent
\textbf{Enhancement Layer:}
The enhancement layer further enriches the representation by applying random nonlinear transformations to the feature output \(Z\). Specifically, \(s\) enhancement windows are generated, each with \(r\) nodes:
\begin{align}
H_{e_j} = \psi(Z W_j + B_j),
\end{align}
where \(W_j \in \mathbb{R}^{pq \times r}\), \(B_j \in \mathbb{R}^{m \times r}\), and \(\psi(\cdot)\) is a nonlinear activation function. The overall enhancement output is
\begin{align}
H = [H_{e_1}, H_{e_2}, \ldots, H_{e_s}] \in \mathbb{R}^{m \times rs}.
\end{align}

\noindent
\textbf{Output Layer:}
The final representation matrix is obtained by concatenating the feature and enhancement outputs:
\begin{align}
A = [Z, H] \in \mathbb{R}^{m \times (pq+rs)}.
\end{align}
BLS determines the output weight matrix \(W \in \mathbb{R}^{(pq+rs) \times n_{\text{out}}}\) by solving a regularized least squares problem:
\begin{align}
\arg\min_{W}
\;\frac{1}{2}\|W\|_{F}^{2}
+
\frac{C}{2}\|AW - Y\|_{F}^{2},
\end{align}
where \(Y\) is the target matrix and \(C>0\) is the regularization parameter. The closed-form solution is given by
\begin{align}\label{BLS-solution}
W =
\begin{cases}
(A^\top A + \frac{1}{C}I)^{-1}A^\top Y, & (pq+rs) \le m, \\
A^\top (A A^\top + \frac{1}{C}I)^{-1}Y, & m < (pq+rs),
\end{cases}
\end{align}
with \(I\) denoting the identity matrix of appropriate dimension.

\section{Proposed Work} \label{Proposed-work}
In this section, we introduce Wave-BLS, i.e., the broad learning system based on the wave loss function. We first discuss the wave loss function and provide its key characteristics. Subsequently, we detail the formulation and optimization of the proposed Wave-BLS model.
\subsection{Wave Loss Function}
\begin{figure*}[t]
    \centering
    \setlength{\tabcolsep}{2pt}
    \begin{tabular}{cccc}
        \includegraphics[width=0.24\textwidth]{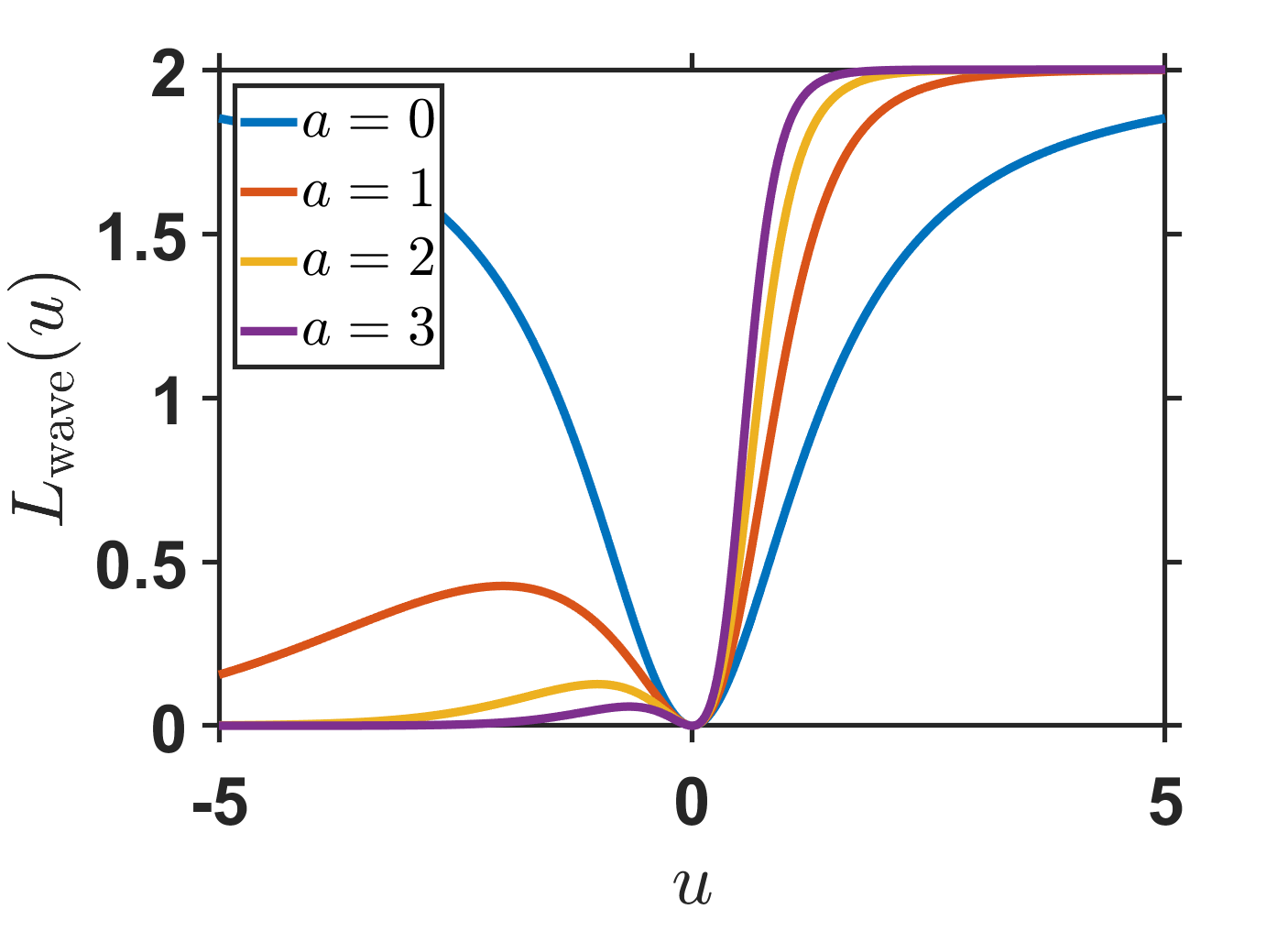} &
        \includegraphics[width=0.24\textwidth]{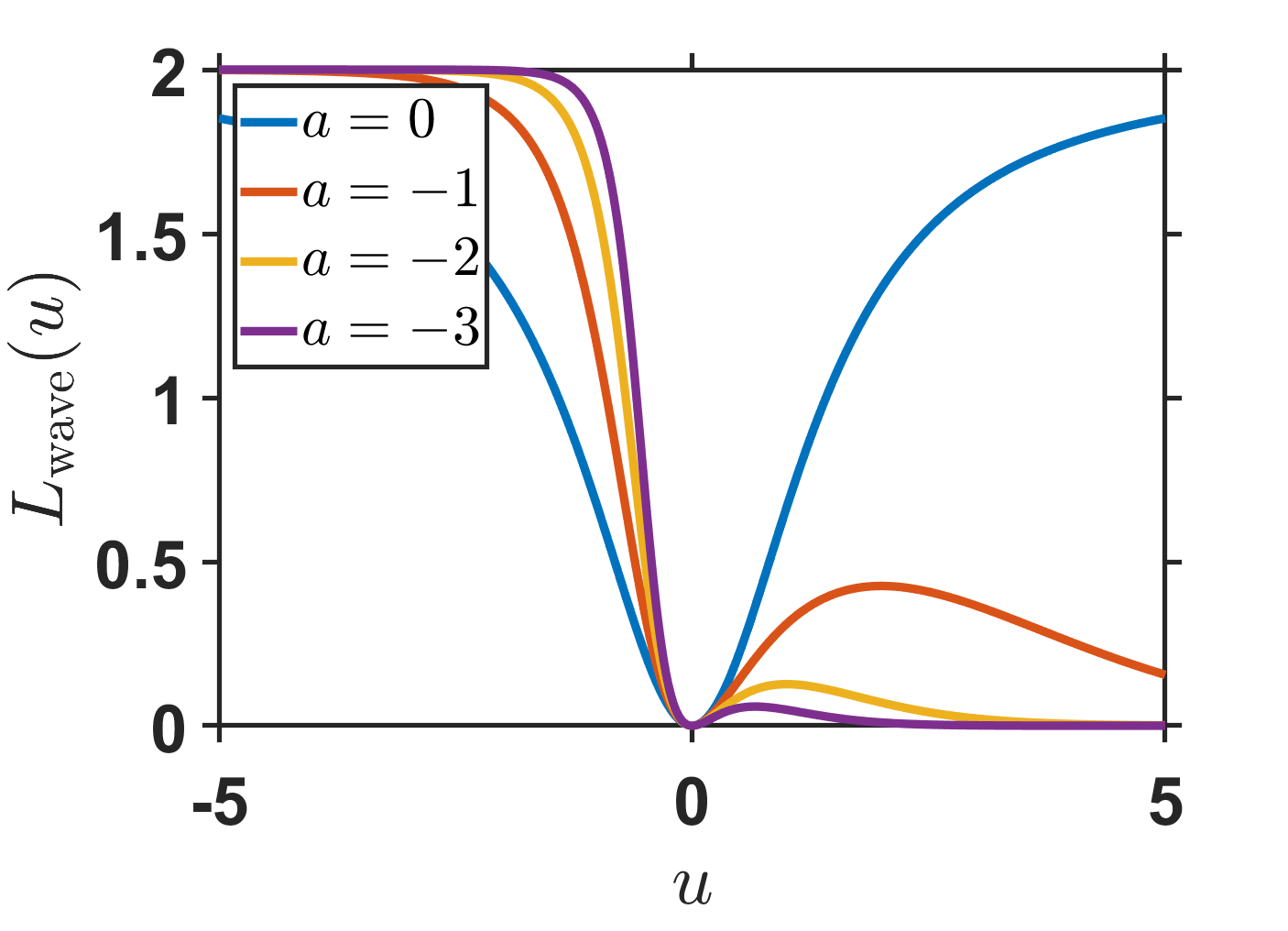} &
        \includegraphics[width=0.24\textwidth]{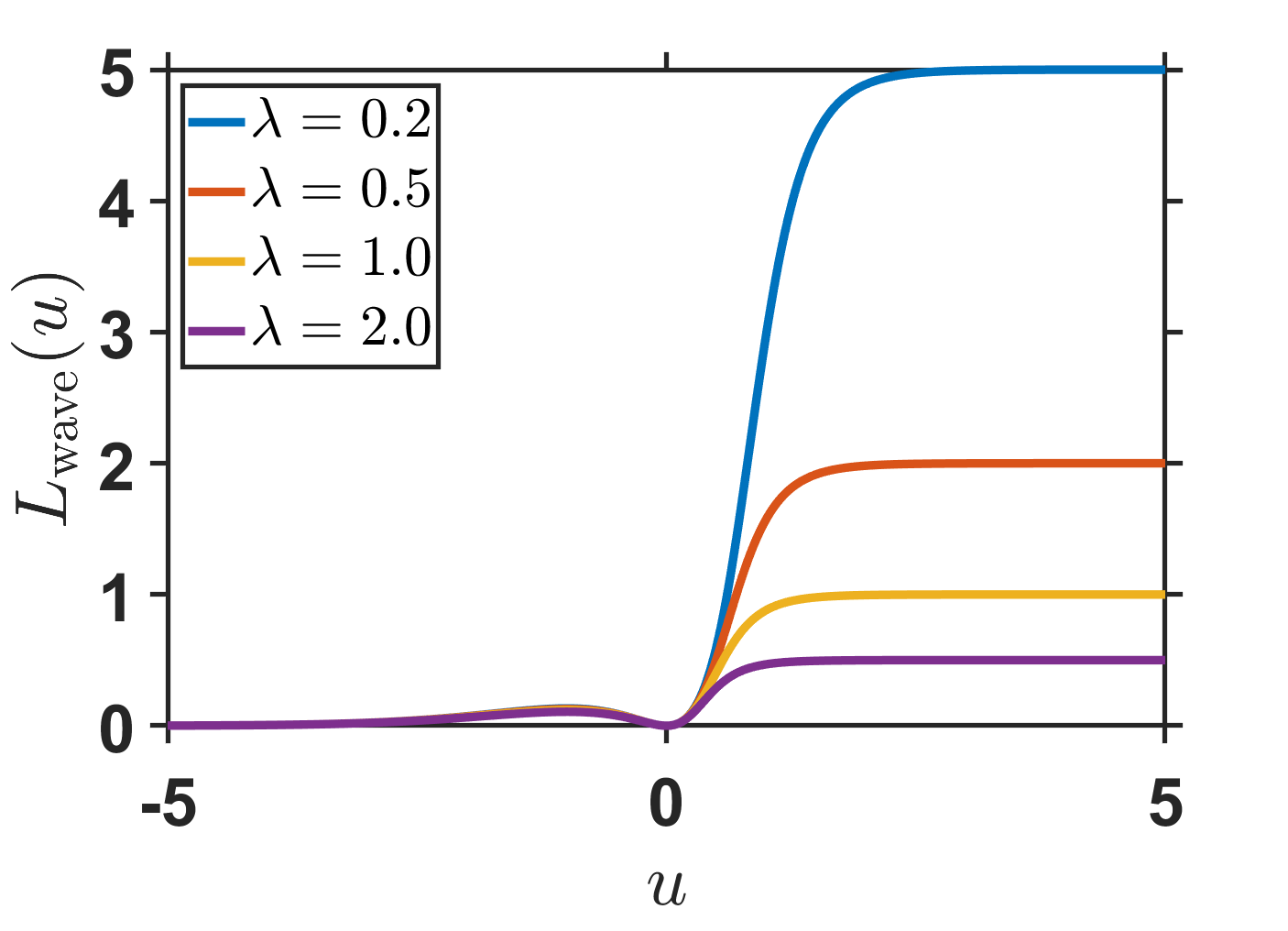} &
        \includegraphics[width=0.24\textwidth]{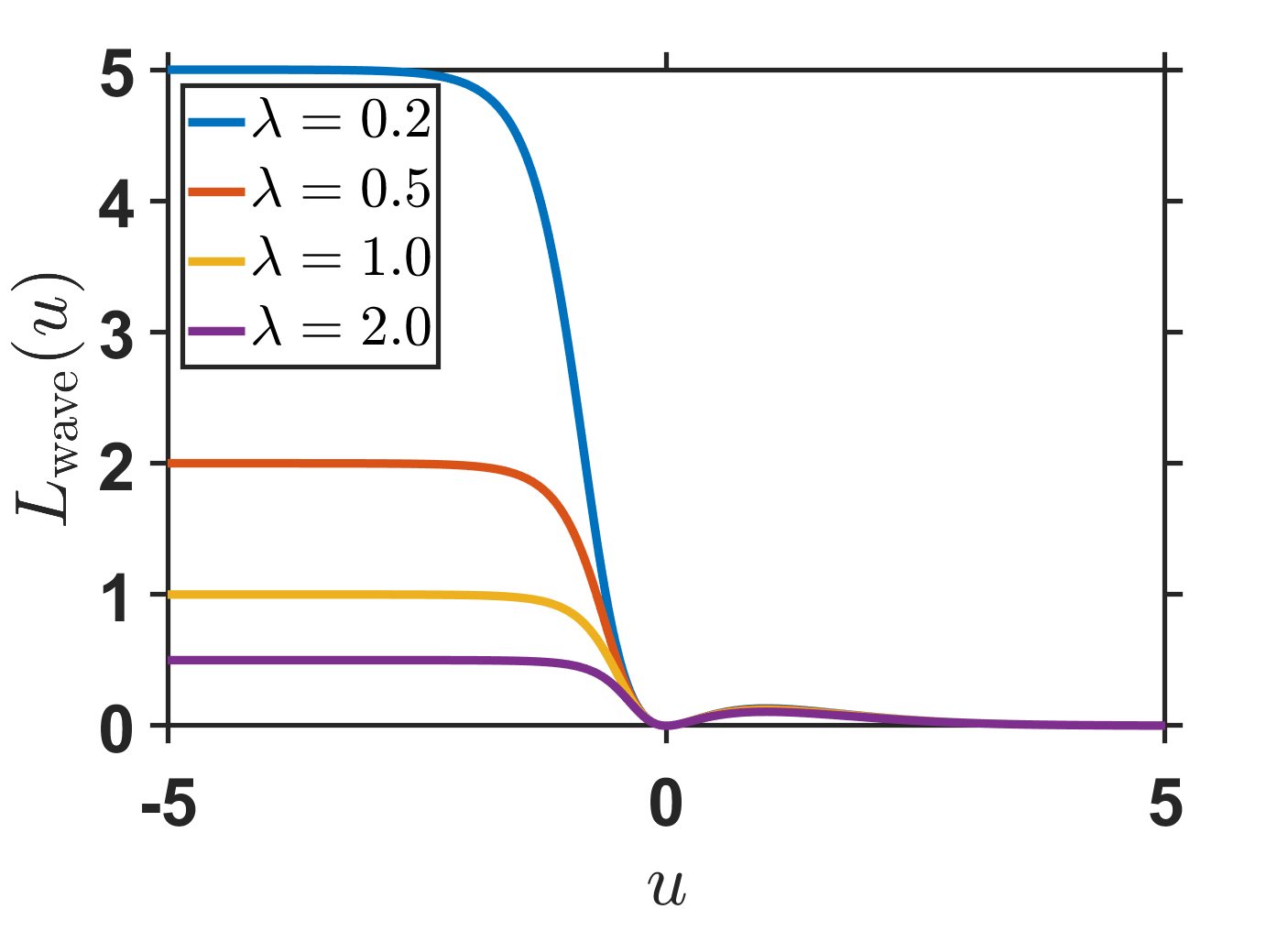}
    \end{tabular}
\caption{Illustration of wave loss function, showing effect of asymmetry parameter \(a\) (positive, negative, and \(a=0\)) and the bounding parameter \(\lambda\).
}
    \label{fig:wave-loss-analysis}
\end{figure*}

The wave loss function \cite{akhtar2024advancing} is a recently introduced asymmetric, bounded, and smooth loss designed to provide controlled error penalization. It is defined as
\begin{align}
L_{\text{wave}}(u)
=
\frac{u^{2} e^{a u}}{1 + \lambda u^{2} e^{a u}},
\end{align}
where \(u \in \mathbb{R}\) denotes the prediction error, \(a \in \mathbb{R}\) is a shape parameter governing the asymmetry and growth behavior of the loss, and \(\lambda \in \mathbb{R}^{+}\) is a bounding parameter that limits the maximum penalty induced by large errors. Fig.~\ref{fig:wave-loss-analysis} illustrates the behavior of the wave loss function under different values of \(a\) and \(\lambda\).

The key properties of the wave loss function, which are directly leveraged in the proposed Wave-BLS framework, are summarized below.
\begin{itemize}
\item \textbf{Asymmetry:}
The wave loss is asymmetric, enabling unequal penalization of overestimation and underestimation errors through the shape parameter \(a\). Positive values of \(a\) emphasize overpredictions, while negative values accentuate underpredictions; the loss becomes symmetric when \(a=0\).

\item \textbf{Boundedness:}
The wave loss is upper-bounded by \(1/\lambda\), ensuring that large errors exert limited influence on the optimization process. This bounded nature prevents excessive domination by outliers and enhances robustness in noisy learning environments.

\item \textbf{Smoothness:}
The wave loss is continuously differentiable, facilitating efficient and stable optimization. This smoothness enables the use of efficient gradient-based solvers.
\end{itemize}

These properties enable the wave loss function to address the fundamental issues of traditional loss functions.

\subsection{Formulation \& Optimization: Broad Learning System with Wave Loss Function (Wave-BLS)}
In this subsection, we incorporate the wave loss function into the BLS by replacing the conventional squared error loss, resulting in a robust learning framework termed broad learning system with wave loss function, i.e., Wave-BLS. The proposed formulation improves robustness against noise and outliers while enabling scalable optimization by avoiding explicit matrix inversion during parameter estimation.

The optimization problem of Wave-BLS is formulated as
\begin{align}\label{Wave-BLS-Optimization-Problem-1}
J(W)
=
\frac{1}{2}\|W\|_{F}^{2}
+
C\,L_{\text{wave}}(AW - Y),
\end{align}
where \(C > 0\) is the regularization parameter controlling the trade-off between model complexity and empirical risk, and \(L_{\text{wave}}(AW - Y)\) represents the aggregated wave loss over all training samples.

The wave loss for the \(i^{th}\) sample and \(j{th}\) class is defined as
\begin{align}
L_{\text{wave}}(u_{ij})
=
\frac{u_{ij}^{2} e^{a u_{ij}}}{1 + \lambda u_{ij}^{2} e^{a u_{ij}}},
\end{align}
where
\(u_{ij} = (AW)_{ij} - Y_{ij}\) denotes the prediction error for sample \(i\) and class \(j\).

Accordingly, the total wave loss over the training set is computed as
\begin{align}
L_{\text{wave}}(AW - Y)
=
\sum_{i=1}^{m}
\sum_{j=1}^{n_{\text{out}}}
L_{\text{wave}}(u_{ij}).
\end{align}

Hence, the objective function in \eqref{Wave-BLS-Optimization-Problem-1} can be equivalently rewritten as
\begin{align}\label{Wave-BLS-Optimization-Problem-2}
J(W)
=
\frac{1}{2}\|W\|_{F}^{2}
+
C \sum_{i=1}^{m} \sum_{j=1}^{n_{\text{out}}}
L_{\text{wave}}(u_{ij}).
\end{align}
The objective function in \eqref{Wave-BLS-Optimization-Problem-2} consists of two components:
\begin{enumerate}
    \item \textbf{Regularization term} \(\big(\frac{1}{2}\|W\|_{F}^{2}\big)\), which controls model complexity and promotes generalization in accordance with the principle of structural risk minimization.
    \item \textbf{Empirical loss term} \(\big( \sum_{i=1}^{m} \sum_{j=1}^{n_{\text{out}}} L_{\text{wave}}(u_{ij})\big)\), which aggregates the wave loss over all samples and classes, thereby enhancing robustness through its asymmetric, bounded, and smooth characteristics.
\end{enumerate}

To efficiently optimize the objective function in \eqref{Wave-BLS-Optimization-Problem-2}, we exploit its smoothness and use a gradient-based optimization strategy. This approach eliminates the need for explicit matrix inversion, which is a computational bottleneck in conventional BLS, thereby significantly improving scalability.

Specifically, we adopt the Nesterov accelerated gradient (NAG) algorithm due to its fast convergence and stable update behavior. The look-ahead mechanism of NAG enables accurate gradient estimates, while its momentum-based formulation effectively suppresses oscillations in ill-conditioned optimization landscapes.

Further, to ensure stable and efficient convergence, the learning rate is dynamically adjusted using an exponential decay schedule. At iteration \(t\), the learning rate is updated as
$
\mu_{t} = \mu_{0}\, e^{-\eta t},
$
where \(\mu_{0}\) is the initial learning rate and \(\eta\) controls the decay rate. This adaptive strategy balances convergence speed and stability, leading to reliable optimization of the proposed Wave-BLS model.

To optimize the objective function in \eqref{Wave-BLS-Optimization-Problem-2}, we compute its gradient with respect to the output weight matrix \(W\). The resulting gradient consists of contributions from the regularization term and the wave loss term and is used within the NAG-based optimization scheme.

For a given class \(j \in \{1,2,\ldots,n_{\text{out}}\}\), the gradient of the wave loss term with respect to \(W(:,j)\) is given by
\begin{align}\label{Gradient-of-wave-loss-formula}
\nabla_{W(:,j)} L_{\text{wave}}
=
\sum_{i=1}^{m}
\frac{u_{ij}\big(2 + a u_{ij}\big)e^{a u_{ij}}}
{\big(1 + \lambda u_{ij}^{2} e^{a u_{ij}}\big)^{2}}
\, A(i,:)^{\top}.
\end{align}
This gradient is subsequently used to update the model parameters within the NAG optimization framework.



The overall gradient of the objective function \(J(W)\) is given by
\begin{align}\label{Gradient-Objective-Function}
\nabla J(W)
=
W
+
C\,\nabla_W L_{\text{wave}},
\end{align}
where the first term arises from the regularization component \(\frac{1}{2}\|W\|_{F}^{2}\), and the second term corresponds to the contribution of the wave loss.

At the \(t^{th}\) iteration, the gradient is computed using the entire training dataset, ensuring stable and deterministic updates. Algorithm~\ref{alg:NAG_Wave-BLS} provides the complete procedure for computing the output weight matrix \(W\) in the proposed Wave-BLS model using the NAG-based optimization scheme. 

The computational complexity of the proposed Wave-BLS model is detailed in Section S.I of the supplementary file.

\begin{algorithm}[t]
\caption{NAG-based optimization for Wave-BLS}
\label{alg:NAG_Wave-BLS}
\textbf{Input:}  
\(A \in \mathbb{R}^{m \times (pq+rs)}\): representation matrix,  
\(Y \in \mathbb{R}^{m \times n_{\text{out}}}\): target matrix, \(C\): regularization parameter, \(I_{\max}\): maximum number of iterations,
\(\mu^{(0)}\): initial learning rate, \(\gamma\): momentum coefficient,  
\(\eta\): learning rate decay factor,  
\(\delta\): convergence tolerance.

\textbf{Output:}  
\(W \in \mathbb{R}^{(pq+rs) \times n_{\text{out}}}\): output weight matrix.

\begin{algorithmic}[1]
\State Initialize \(t \leftarrow 0\), \(W^{(0)}\), \(v^{(0)} \leftarrow \mathbf{0}\), \(\mu^{(0)}\).
\Repeat
    \State Compute:
    $W^{\text{look-ahead}} = W^{(t)} + \gamma v^{(t)}$.
    \State Compute stochastic gradient \(g^{(t)} = \nabla J(W^{\text{look-ahead}})\).
    \State Velocity update:
    \[
    v^{(t+1)} = \gamma v^{(t)} - \mu^{(t)} g^{(t)}.
    \]
    \State Model parameters update:
    \[
    W^{(t+1)} = W^{(t)} + v^{(t+1)}.
    \]
    \State Learning rate update:
    \[
    \mu^{(t+1)} = \mu^{(t)} e^{-\eta t}.
    \]
    \State \(t \leftarrow t + 1\).
\Until{\(t \geq I_{\max}\) \textbf{or} \(\|W^{(t)} - W^{(t-1)}\|_{F} \leq \delta\)}
\State \Return \(W\).
\end{algorithmic}
\end{algorithm}

\section{Experiments and Discussion} \label{Experiment-section}
This section empirically evaluates the proposed Wave-BLS against baseline models on 30 UCI benchmark datasets under both clean and noisy conditions. The proposed Wave-BLS model is evaluated against several baseline methods, including RVFL \cite{pao1994learning}, ELM, i.e., RVFL without direct link(RVFLwoDL) \cite{huang2006extreme}, BLS \cite{chen2017broad2}, Wave-RVFL \cite{sajid2024wavervflrandomizedneuralnetwork}, NF-BLS \cite{feng2018fuzzy}, F-BLS \cite{sajid2024intuitionistic}, IF-BLS \cite{sajid2024intuitionistic}, and KRP-BLS \cite{10902561}. The detailed experimental setup is provided in Section S.II of the supplementary material.

\begin{table*}[t]
\centering
\renewcommand{\arraystretch}{1.2}
\caption{\small{Comparison of average classification performance of the proposed Wave-BLS and baseline models 30 UCI benchmark datasets.}}
\label{tab:acc_rank_comparison}
\resizebox{\textwidth}{!}{
\begin{tabular}{lccccccccc}
\toprule
 \textbf{Model} $\rightarrow$ & \textbf{RVFL \cite{pao1994learning}} & \textbf{RVFLwoDL \cite{huang2006extreme}} & \textbf{BLS \cite{chen2017broad2}} & \textbf{Wave-RVFL \cite{sajid2024wavervflrandomizedneuralnetwork}} & \textbf{NF-BLS \cite{feng2018fuzzy}} & \textbf{F-BLS \cite{sajid2024intuitionistic}} & \textbf{IF-BLS \cite{sajid2024intuitionistic}} & \textbf{KRP-BLS \cite{10902561}} & \textbf{Wave-BLS$^{\dagger}$} \\
\midrule
\textbf{Average Accuracy} & 79.8876 & 79.4376 & 82.0751 & 81.5155 & 80.0399 & 82.0376 & \underline{83.9102} & 83.002 & \textbf{86.7448} \\
\textbf{Average Rank} & 6.5333 & 7.1833 & 4.8 & 4.0667 & 6.0833 & 5.2667 & \underline{3.4833} & 4.8333 & \textbf{2.75} \\
\bottomrule
\multicolumn{10}{l}{$^{\dagger}$ indicates the proposed model; \textbf{bold} denotes the best performance, and \underline{underline} denotes the second-best.}
\end{tabular}
}
\end{table*}

\subsection{Performance Evaluation}
Table~\ref{tab:acc_rank_comparison} summarizes the average accuracy and average rank of the proposed Wave-BLS and competing methods over 30 UCI benchmark datasets. The detailed results are provided in Table S.I of the supplement file. Overall, Wave-BLS achieves the best performance in both criteria, indicating not only higher accuracy but also more consistent dominance across datasets.
Among the classical randomized baselines, RVFL and RVFLwoDL obtain average accuracies of 79.8876\% and 79.4376\%, respectively. Moving to the broad-learning family, the standard BLS improves the average accuracy to 82.0751\%. Several robust BLS variants further enhance performance: NF-BLS achieves 80.0399\%, F-BLS reaches 82.0376\%, IF-BLS attains the second-best average accuracy of 83.9102\%, and KRP-BLS achieves 83.002\%. In contrast, the proposed Wave-BLS yields the highest average accuracy of 86.7448\%, providing a clear improvement over the strongest baselines. Specifically, Wave-BLS improves upon the best-performing baseline IF-BLS by 2.8346\% (86.7448\% vs.\ 83.9102\%) and over KRP-BLS by 3.7428\% (86.7448\% vs.\ 83.002\%). Moreover, comparing wave-loss-based models highlights the benefit of embedding wave loss into the broader BLS architecture: Wave-BLS outperforms Wave-RVFL by a large margin of 5.2293\% (86.7448\% vs.\ 81.5155\%), indicating that the proposed formulation leverages both the robustness of wave loss and the stronger representation capacity of BLS. While average accuracy provides a useful aggregate summary, it can hide dataset-wise variability: a method may perform extremely well on a subset of datasets and poorly on others while still achieving a competitive mean. Average rank complements this by measuring relative performance per dataset and then aggregating, thereby better reflecting how consistently a method performs across heterogeneous benchmarks. Lower average rank indicates that a model attains top performance more frequently across datasets, rather than relying on a few large gains. The rank results in Table~\ref{tab:acc_rank_comparison} further confirm the advantage of Wave-BLS. RVFL and RVFLwoDL obtain average ranks of 6.5333 and 7.1833, respectively, whereas the standard BLS improves to 4.8. Among robust BLS variants, IF-BLS achieves the second-best average rank of 3.4833, while KRP-BLS yields 4.833. Wave-BLS achieves the best average rank of 2.75, demonstrating the most consistent superiority across datasets. Quantitatively, Wave-BLS improves over IF-BLS by 0.7333 (3.4833 $\rightarrow$ 2.75), and over KRP-BLS by 2.0833 (4.8333 $\rightarrow$ 2.75). A similar trend is observed when comparing the wave-loss counterparts: Wave-BLS significantly improves over Wave-RVFL in rank (4.0667 $\rightarrow$ 2.75), confirming that the proposed integration of wave loss within the BLS framework not only boosts mean accuracy but also yields more consistent dataset-wise wins. Taken together, the lowest average rank and highest average accuracy indicate that Wave-BLS delivers both stronger predictive performance and better cross-dataset stability than existing baselines. This advantage is primarily driven by (i) replacing the squared-error criterion with the robust wave loss and (ii) coupling it with the richer feature-enhancement representation of BLS, yielding substantial gains over both the strongest BLS competitors (IF-BLS, KRP-BLS) and the wave-loss-based RVFL baseline (Wave-RVFL).

To assess whether the observed differences in average ranks among the competing models are statistically significant, we employ the nonparametric Friedman test followed by Nemenyi post-hoc test. The detailed results are provided in Section S.III of the supplementary. The sensitivity analysis of the wave loss hyperparameters for the proposed Wave-BLS model is reported in Section S.IV of the supplementary material.

\begin{table}[t]
\centering
\renewcommand{\arraystretch}{1.15}
\caption{Robustness comparison of BLS and Wave-BLS under outlier and noise contamination.}
\label{tab:robustness_wave_bls}
\resizebox{0.95\columnwidth}{!}{
\begin{tabular}{|l|c|cc|c|cc|}
\hline
\multirow{2}{*}{Dataset} & \multirow{2}{*}{Level}
& \multicolumn{2}{c|}{Outliers}
& \multirow{2}{*}{Level}
& \multicolumn{2}{c|}{Noise} \\ \cline{3-4} \cline{6-7}
 &  & BLS & Wave-BLS$^{\dagger}$ &  & BLS & Wave-BLS$^{\dagger}$ \\ \hline

\multirow{6}{*}{blood}
& Clean & 79.6272 & \textbf{82.9184} & Clean & 79.6272 & \textbf{82.9184} \\
& 5\%   & 75.8421 & \textbf{80.4067} & 5\%   & 77.1034 & \textbf{81.7749} \\
& 10\%  & 72.5186 & \textbf{78.3925} & 10\%  & 74.8652 & \textbf{79.9368} \\
& 15\%  & 69.2847 & \textbf{75.8106} & 15\%  & 71.4928 & \textbf{77.1053} \\
& 20\%  & 66.0319 & \textbf{72.4678} & 20\%  & 68.1074 & \textbf{74.6389} \\ \cline{2-7}
& Avg.  & 72.6609 & \textbf{78.3992} & Avg.  & 74.3398 & \textbf{79.2749} \\ \hline

\multirow{6}{*}{horse\_colic}
& Clean & \textbf{86.1422} & 85.7919 & Clean & \textbf{86.1422} & 85.7919 \\
& 5\%   & 82.6374 & \textbf{84.9563} & 5\%   & 83.9046 & \textbf{85.1187} \\
& 10\%  & 78.9421 & \textbf{82.4018} & 10\%  & 80.2873 & \textbf{82.9365} \\
& 15\%  & 75.6189 & \textbf{79.6674} & 15\%  & 76.9548 & \textbf{80.3412} \\
& 20\%  & 72.3046 & \textbf{76.1897} & 20\%  & 73.6285 & \textbf{77.5403} \\ \cline{2-7}
& Avg.  & 79.1290 & \textbf{81.8014} & Avg.  & 80.1835 & \textbf{82.3457} \\ \hline

\multicolumn{2}{|l|}{Overall Average}
& 75.8949 & \textbf{80.1003}
&  & 77.2617 & \textbf{80.8103} \\ \hline

\multicolumn{2}{|l|}{Average Rank}
& 2.0 & \textbf{1.0}
&  & 1.75 & \textbf{1.25} \\ \hline

\end{tabular}
}
\vspace{1mm}
{\footnotesize \textbf{Bold} values denote the best performance for each row.}
\end{table}
\subsection{Robustness Analysis under Noise and Outliers}
To evaluate the robustness of the proposed Wave-BLS, we conduct controlled experiments under outlier and noise contamination on two representative datasets. Specifically, artificial perturbations are introduced by contaminating $5\%$, $10\%$, $15\%$, and $20\%$ of the training samples with outliers and label noise, respectively. The blood dataset is selected as a case where Wave-BLS consistently outperforms the standard BLS, while the horse\_colic dataset represents a complementary scenario where BLS exhibits slightly better performance on the clean data. This selection enables a balanced and unbiased assessment of robustness behavior. As reported in Table~\ref{tab:robustness_wave_bls}, Wave-BLS demonstrates clear robustness advantages on the blood dataset across all contamination levels. Under both outlier and noise perturbations, Wave-BLS maintains higher classification accuracy than BLS, with performance degradation occurring at a noticeably slower rate as contamination increases. This behavior highlights the effectiveness of the wave loss in bounding the influence of corrupted samples and mitigating the impact of extreme deviations. For the horse\_colic dataset, although BLS achieves marginally higher accuracy on the clean data, Wave-BLS consistently surpasses BLS once outliers or noise are introduced. As contamination levels increase, Wave-BLS preserves stronger performance stability, leading to higher average accuracy under both outlier and noise settings. This indicates that even in cases where the baseline model performs well on clean data, the proposed wave loss enhances resilience under adverse conditions. Overall, Wave-BLS attains superior average accuracy and lower average rank across both datasets and contamination types. These results confirm that the proposed Wave-BLS framework offers improved robustness to data corruption, making it more suitable for real-world learning scenarios where noise and outliers are unavoidable.

\section{Conclusions} \label{Conclusions-section}
In this work, we proposed Wave-BLS, a robust Broad Learning System that integrates the wave loss function into the BLS framework to address its inherent sensitivity to noise and outliers. By replacing the least-squares error formulation with a boundedand and asymmetric loss and optimizing the resulting objective via NAG-based scheme, the proposed approach eliminates the need for matrix inversion while preserving the flat and efficient architecture of BLS. Extensive experiments on several benchmark datasets show that Wave-BLS achieves consistently higher classification performance than baseline variants. The superiority of the proposed model is further supported by Friedman and Nemenyi statistical tests. These results indicate that replacing the squared-error criterion with a bounded loss can substantially improve the reliability of BLS-based models. Robustness studies conducted under controlled noise and outlier injection provide additional insight into the behavior of Wave-BLS in adverse learning conditions. While standard BLS may perform competitively on clean data, its performance degrades rapidly as contamination increases. Moreover, sensitivity analysis with respect to the wave loss hyperparameters reveals a wide region of stable performance, indicating that Wave-BLS does not depend on finely tuned parameter settings. Overall, the empirical evidence suggests that Wave-BLS offers a robust enhancement to the BLS framework without introducing architectural complexity.

\bibliographystyle{IEEEtranN}
\bibliography{refs.bib}

\clearpage
\setcounter{section}{0}
\setcounter{subsection}{0}
\setcounter{table}{0}
\setcounter{figure}{0}
\setcounter{equation}{0}
\renewcommand{\thesection}{S.\Roman{section}}
\renewcommand{\thesubsection}{\thesection.\Alph{subsection}}
\renewcommand{\thetable}{S.\Roman{table}}
\renewcommand{\thefigure}{S.\arabic{figure}}
\renewcommand{\theequation}{S.\arabic{equation}}
\renewcommand{\theHsection}{supp.\Roman{section}}
\renewcommand{\theHtable}{supp.\Roman{table}}
\renewcommand{\theHfigure}{supp.\arabic{figure}}
\renewcommand{\theHequation}{supp.\arabic{equation}}
\twocolumn[
\begin{center}
{\LARGE\bfseries Supplementary Material}\\
{\large Robust Broad Learning System with Wave Loss for Classification under Data Uncertainty}
\end{center}
\vspace{1ex}
]

\section{Computational Complexity}
The computational complexity of the proposed Wave-BLS model optimized using the NAG framework is mainly determined by the gradient computation and parameter update steps. At each iteration, the gradient of the objective function \(\nabla J(W)\) is evaluated over the entire training set, requiring the processing of all \(m\) samples across \(n_{\text{out}}\) output dimensions. For each sample, the computation involves a scalar weighting followed by a multiplication with the transformed feature vector \(A(i,:)\in\mathbb{R}^{(pq+rs)}\), resulting in a per-iteration complexity of \(\mathcal{O}\big(m\,n_{\text{out}}\, (pq+rs)\big)\). The subsequent NAG operations, including the look-ahead step, velocity update, and parameter update, operate on matrices of size \((pq+rs)\times n_{\text{out}}\) and therefore contribute a lower-order cost of \(\mathcal{O}\big((pq+rs)\,n_{\text{out}}\big)\). Since the gradient computation dominates, the overall per-iteration complexity remains \(\mathcal{O}\big(m\,n_{\text{out}}\, (pq+rs)\big)\). Over \(I_{\max}\) iterations, the total computational complexity of Wave-BLS is \(\mathcal{O}\big(I_{\max}\, m\, n_{\text{out}}\, (pq+rs)\big)\). In contrast, the conventional BLS framework requires matrix inversion with a computational cost of \(\mathcal{O}\!\left(\min\!\left((pq+rs)^3,\, m^3\right)\right)\), highlighting the improved scalability of the proposed Wave-BLS model.

\section{Experimental Setup}
All experiments are conducted using MATLAB R2023a on a workstation equipped with an 11th-generation Intel Core i7-11700 processor running at 2.50~GHz, 16~GB of RAM, and a Windows~11 operating system. To ensure a fair and reliable comparison, a 5-fold cross-validation strategy is employed in conjunction with grid search for hyperparameter tuning. Specifically, each dataset is partitioned into five mutually exclusive folds. For a given set of hyperparameters, the model is trained on four folds and evaluated on the remaining fold, repeating this process until each fold has served as the test set once. The testing accuracy obtained across the five folds is then averaged, and the highest average accuracy achieved over all hyperparameter configurations is reported as the final performance of each model.

For all models, the regularization parameter \(C\) is selected from the range \(\{10^{-6}, 10^{-4}, \ldots, 10^{6}\}\). For RVFL and ELM, the number of hidden nodes is tuned within the range \(5{:}10{:}205\). For BLS and the proposed Wave-BLS, the number of feature node windows is chosen from \(1{:}2{:}21\), the number of feature nodes per window from \(5{:}5{:}50\), and the number of enhancement nodes from \(5{:}10{:}105\). For NF-BLS, the number of fuzzy groups is selected from \(1{:}2{:}21\), the number of fuzzy nodes per group from \(5{:}5{:}50\), and the number of enhancement nodes from \(5{:}10{:}105\). For IF-RVFL, the kernel hyperparameter is selected from the range \(\{10^{-6}, 10^{-4}, \ldots, 10^{6}\}\). For Wave-RVFL, hyperparameter settings are adopted from \cite{sajid2024wavervflrandomizedneuralnetwork} to ensure a fair and consistent comparison. Similarly, the configurations for F-BLS and IF-BLS follow the settings reported in \cite{sajid2024intuitionistic}, while the parameters for KRP-BLS are set according to \cite{10902561}, maintaining consistency with their original implementations. For the proposed Wave-BLS model, the wave loss parameters are selected from \(a \in \{-5, -4, \ldots, 5\}\) and \(\lambda \in \{0.1, 0.5, 1\}\). The parameters of the NAG-based optimization algorithm are fixed as follows: the initial weight matrix is set to \(W^{(0)} = 0.01 \times \mathbf{1}\), where \(\mathbf{1} \in \mathbb{R}^{(pq+rs)\times n_{\text{class}}}\) denotes a matrix of ones; the initial velocity is initialized as \(v^{(0)} = \mathbf{0}\); the initial learning rate is set to \(\mu^{(0)} = 0.01\); the learning rate decay factor is fixed to \(\eta = 0.1\); the momentum coefficient is set to \(\gamma = 0.6\); the convergence tolerance is set to \(\delta = 10^{-6}\); and the maximum number of iterations is fixed to \(I_{\max} = 100\).

\begin{table*}[t]
\centering
\renewcommand{\arraystretch}{1.2}
\caption{Dataset wise results of the proposed wave-BLS model against the baseline models on 30 UCI datasets.}
\label{tab:supp_acc_rank_comparison}
\resizebox{\textwidth}{!}{
\begin{tabular}{lccccccccc}
\toprule
\textbf{Dataset} $\downarrow$ $|$ \textbf{Model} $\rightarrow$ & \textbf{RVFL \cite{pao1994learning}} & \textbf{RVFLwoDL \cite{huang2006extreme}} & \textbf{BLS \cite{chen2017broad2}} & \textbf{Wave-RVFL \cite{sajid2024wavervflrandomizedneuralnetwork}} & \textbf{NF-BLS \cite{feng2018fuzzy}} & \textbf{F-BLS \cite{sajid2024intuitionistic}} & \textbf{IF-BLS \cite{sajid2024intuitionistic}} & \textbf{KRP-BLS \cite{10902561}} & \textbf{Wave-BLS$^{\dagger}$} \\
\midrule

acute\_inflammation & 100 & 100 & 100 & 100 & 100 & 100 & 100 & 100 & 100 \\
acute\_nephritis & 100 & 100 & 100 & 100 & 100 & 100 & 100 & 100 & 100 \\
bank & 90.9051 & 89.4051 & 89.7366 & 88.1779 & 89.5817 & 89.7590 & 89.4051 & 87.7230 & 89.0202 \\
blood & 78.4056 & 76.9056 & 79.6272 & 80.7578 & 76.0807 & 78.5807 & 77.2451 & 76.7000 & 82.9184 \\
breast\_cancer & 68.1727 & 66.6727 & 69.8851 & 70.2179 & 70.1754 & 72.2868 & 83.1579 & 82.4947 & 90.8772 \\
breast\_cancer\_wisc & 89.4897 & 87.9897 & 88.4183 & 86.8050 & 90.7081 & 88.2775 & 88.9938 & 89.3240 & 97.8571 \\
chess\_krvkp & 73.5312 & 72.0312 & 84.3862 & 75.7371 & 70.4004 & 84.2921 & 84.9184 & 87.3718 & 88.7031 \\
conn\_bench\_sonar\_mines\_rocks & 62.0807 & 60.5807 & 69.2451 & 63.9431 & 60.6272 & 69.1521 & 80.2091 & 78.2049 & 93.6585 \\
credit\_approval & 86.8623 & 85.3623 & 87.5362 & 88.5995 & 84.4928 & 86.3768 & 88.5507 & 89.7797 & 87.1014 \\
cylinder\_bands & 67.7193 & 66.2193 & 69.9258 & 69.7509 & 69.5336 & 69.1338 & 72.8536 & 70.8109 & 70.3008 \\
echocardiogram & 85.4031 & 83.9031 & 83.9316 & 87.9652 & 80.9402 & 84.6724 & 88.4900 & 86.7202 & 86.2108 \\
fertility & 89.5000 & 90.0000 & 90.0000 & 86.8150 & 92.0000 & 91.0000 & 91.0000 & 89.1800 & 92.0000 \\
haberman\_survival & 74.9902 & 73.4902 & 70.2750 & 77.2399 & 73.4902 & 69.6563 & 75.4574 & 76.1937 & 78.7361 \\
hepatitis & 83.2258 & 83.2258 & 85.1613 & 85.7226 & 87.7419 & 84.5161 & 87.7419 & 85.1096 & 85.2258 \\
hill\_valley & 77.9706 & 77.9706 & 82.0185 & 80.3097 & 78.9583 & 81.6002 & 79.6249 & 78.6344 & 80.4570 \\
horse\_colic & 84.2098 & 84.7098 & 86.1422 & 86.7361 & 83.9726 & 86.1385 & 86.6938 & 84.0920 & 85.7919 \\
mammographic & 79.0889 & 79.0889 & 78.6712 & 81.4616 & 79.5040 & 78.6744 & 79.8192 & 79.8238 & 84.2870 \\
molec\_biol\_promoter & 68.9610 & 68.9610 & 83.9394 & 71.0298 & 75.1515 & 84.9351 & 88.7879 & 89.1241 & 96.1905 \\
monks\_1 & 83.2497 & 83.2497 & 75.3314 & 85.7472 & 86.1277 & 77.1396 & 77.6705 & 74.0404 & 74.1203 \\
musk\_1 & 67.8640 & 67.8640 & 75.8311 & 69.8999 & 67.8487 & 76.6754 & 78.9846 & 79.8523 & 96.6316 \\
oocytes\_merluccius\_nucleus\_4d & 79.8407 & 79.8407 & 82.7776 & 82.2359 & 80.1420 & 81.8011 & 80.9201 & 79.3017 & 72.8541 \\
oocytes\_trisopterus\_nucleus\_2f & 76.4211 & 76.4211 & 78.9371 & 78.7137 & 75.9911 & 78.6123 & 75.9899 & 74.2302 & 76.2815 \\
pima & 71.4888 & 71.4888 & 72.0958 & 73.6335 & 72.7918 & 71.4846 & 73.3079 & 74.1087 & 76.4341 \\
pittsburg\_bridges\_T\_OR\_D & 87.6429 & 88.1429 & 89.1095 & 90.2722 & 90.1429 & 88.1905 & 90.2381 & 87.0791 & 92.1429 \\
spambase & 87.1125 & 87.1125 & 89.0201 & 89.7259 & 83.2657 & 89.4582 & 90.7407 & 87.4741 & 93.9201 \\
spect & 67.9245 & 67.9245 & 69.4340 & 69.9622 & 67.9245 & 69.4340 & 72.4528 & 72.3517 & 73.9623 \\
statlog\_heart & 80.0000 & 80.0000 & 82.2222 & 82.4000 & 80.7407 & 82.2222 & 84.0741 & 85.0566 & 87.0370 \\
tic\_tac\_toe & 86.0068 & 86.0068 & 98.4315 & 88.5870 & 82.7623 & 97.8081 & 97.0768 & 94.1645 & 99.0201 \\
titanic & 77.9168 & 77.9168 & 77.9168 & 80.2543 & 78.0537 & 77.9168 & 79.9532 & 78.3541 & 94.6875 \\
vertebral\_column\_2clases & 70.6452 & 70.6452 & 72.2453 & 72.7646 & 72.0465 & 71.3333 & 72.9479 & 72.7590 & 75.9168 \\
\midrule
\textbf{Average Accuracy} & 79.8876 & 79.4376 & 82.0751 & 81.5155 & 80.0399 & 82.0376 & \underline{83.9102} & 83.002 & \textbf{86.7448} \\
\textbf{Average Rank} & 6.5333 & 7.1833 & 4.8 & 4.0667 & 6.0833 & 5.2667 & \underline{3.4833} & 4.8333 & \textbf{2.75} \\
\bottomrule
\multicolumn{10}{l}{$^{\dagger}$ indicates the proposed model; \textbf{bold} denotes the best performance, and \underline{underline} denotes the second-best.}
\end{tabular}
}
\end{table*}

\section{Statistical Significance Analysis}
To assess whether the observed differences in average ranks among the competing models are statistically significant, we first employ the nonparametric Friedman test. Let $\mathcal{K}$ denote the number of evaluated models and $\mathcal{N}$ denote the number of datasets. This test examines the null hypothesis that all $\mathcal{K}$ models exhibit equivalent performance across $\mathcal{N}$ datasets. The Friedman test statistic is defined as
$
\chi^2_F
=
\frac{12\mathcal{N}}{\mathcal{K}(\mathcal{K}+1)}
\left(
\sum_{k=1}^{\mathcal{K}} \varrho(k,\bullet)^2
-
\frac{\mathcal{K}(\mathcal{K}+1)^2}{4}
\right),
$
where $\varrho(k,\bullet)$ denotes the average rank of the $k^{th}$ model. In the test,
the statistic
$
F_F
=
\chi^2_F
\left(
\frac{\mathcal{N}-1}{\mathcal{N}(\mathcal{K}-1)-\chi^2_F}
\right)
$
is commonly used, which follows an $F$-distribution with $(\mathcal{K}-1)$ and $(\mathcal{K}-1)(\mathcal{N}-1)$ degrees of freedom. In our study, $\mathcal{K}=9$ models are evaluated on $\mathcal{N}=30$ UCI benchmark datasets. As reported in Table~\ref{tab:statistical_tests}, the Friedman statistic is $\chi^2_F = 66.6578$, yielding a corrected value of $F_F = 11.1518$. The corresponding critical value at the $5\%$ significance level is $F(8,232)=1.9784$. As the computed statistic satisfies \(11.1518 > 1.9784\), the null hypothesis of equivalent model performance is rejected, confirming the presence of statistically significant differences across the evaluated methods. Next, the Nemenyi post-hoc procedure is employed to conduct pairwise comparisons between the proposed Wave-BLS and the competing models. According to the Nemenyi test, two models are considered significantly different if the difference in their average ranks exceeds the critical difference (C.D.), given by
$
\text{C.D.}
=
q_{\alpha}
\sqrt{\frac{\mathcal{K}(\mathcal{K}+1)}{6\mathcal{N}}},
$
where $q_{\alpha}$ is the critical value corresponding to the two-tailed Nemenyi test. At $\alpha = 0.1$, the resulting critical difference is $\text{C.D.} = 2.0188$. The post-hoc comparison results are summarized in Table~\ref{tab:statistical_tests}. The proposed Wave-BLS achieves the best average rank of $2.75$ an d shows statistically significant improvements over RVFL, RVFLwoDL, BLS, NF-BLS, F-BLS, and KRP-BLS, as the corresponding rank differences exceed the critical difference. In contrast, the rank differences between Wave-BLS and Wave-RVFL, as well as IF-BLS, do not exceed the critical threshold, indicating comparable performance among these top-performing methods. Overall, the Friedman and Nemenyi tests jointly confirm that the superior performance of Wave-BLS is not due to random variation, but is statistically significant across the considered benchmark datasets.

\begin{table}[t]
\centering
\caption{\small{Statistical significance analysis using Friedman test and Nemenyi post-hoc test.}}
\label{tab:statistical_tests}
\resizebox{9cm}{!}{%
\begin{tabular}{|l|c|c|c|c|c|}
\hline
\multicolumn{6}{|c|}{\textbf{Friedman Test}} \\ \hline
$\mathcal{K}$ 
& $\mathcal{N}$ 
& $\chi^2_F$ 
& $F_F$ 
& $F((\mathcal{K}-1),(\mathcal{K}-1)(\mathcal{N}-1))$ 
& \begin{tabular}[c]{@{}c@{}}Significant\\ difference\end{tabular} \\ \hline
9 & 30 & 66.6578 & 11.1518 & 1.9784 & Yes \\ \hline
\multicolumn{6}{|c|}{\textbf{Nemenyi Post-hoc Test (C.D.\,=\,2.0188)}} \\ \hline
Model 
& Avg. rank 
& Rank diff. 
& \multicolumn{3}{c|}{Significant difference} \\ \hline
RVFL \cite{pao1994learning} 
& 6.5333 & 3.7833 & \multicolumn{3}{c|}{Yes} \\ \hline
RVFLwoDL \cite{huang2006extreme} 
& 7.1833 & 4.4333 & \multicolumn{3}{c|}{Yes} \\ \hline
BLS \cite{chen2017broad2} 
& 4.8000 & 2.0500 & \multicolumn{3}{c|}{Yes} \\ \hline
Wave-RVFL \cite{sajid2024wavervflrandomizedneuralnetwork} 
& 4.0667 & 1.3167 & \multicolumn{3}{c|}{No} \\ \hline
NF-BLS \cite{feng2018fuzzy} 
& 6.0833 & 3.3333 & \multicolumn{3}{c|}{Yes} \\ \hline
F-BLS \cite{sajid2024intuitionistic} 
& 5.2667 & 2.5167 & \multicolumn{3}{c|}{Yes} \\ \hline
IF-BLS \cite{sajid2024intuitionistic} 
& 3.4833 & 0.7333 & \multicolumn{3}{c|}{No} \\ \hline
KRP-BLS \cite{10902561} 
& 4.8333 & 2.0833 & \multicolumn{3}{c|}{Yes} \\ \hline
Wave-BLS$^{\dagger}$ 
& 2.7500 & -- & \multicolumn{3}{c|}{N/A} \\ \hline
\end{tabular}%
}
\end{table}

\begin{figure*}[t]
    \centering
    \subfloat[]{
        \includegraphics[width=0.30\textwidth]{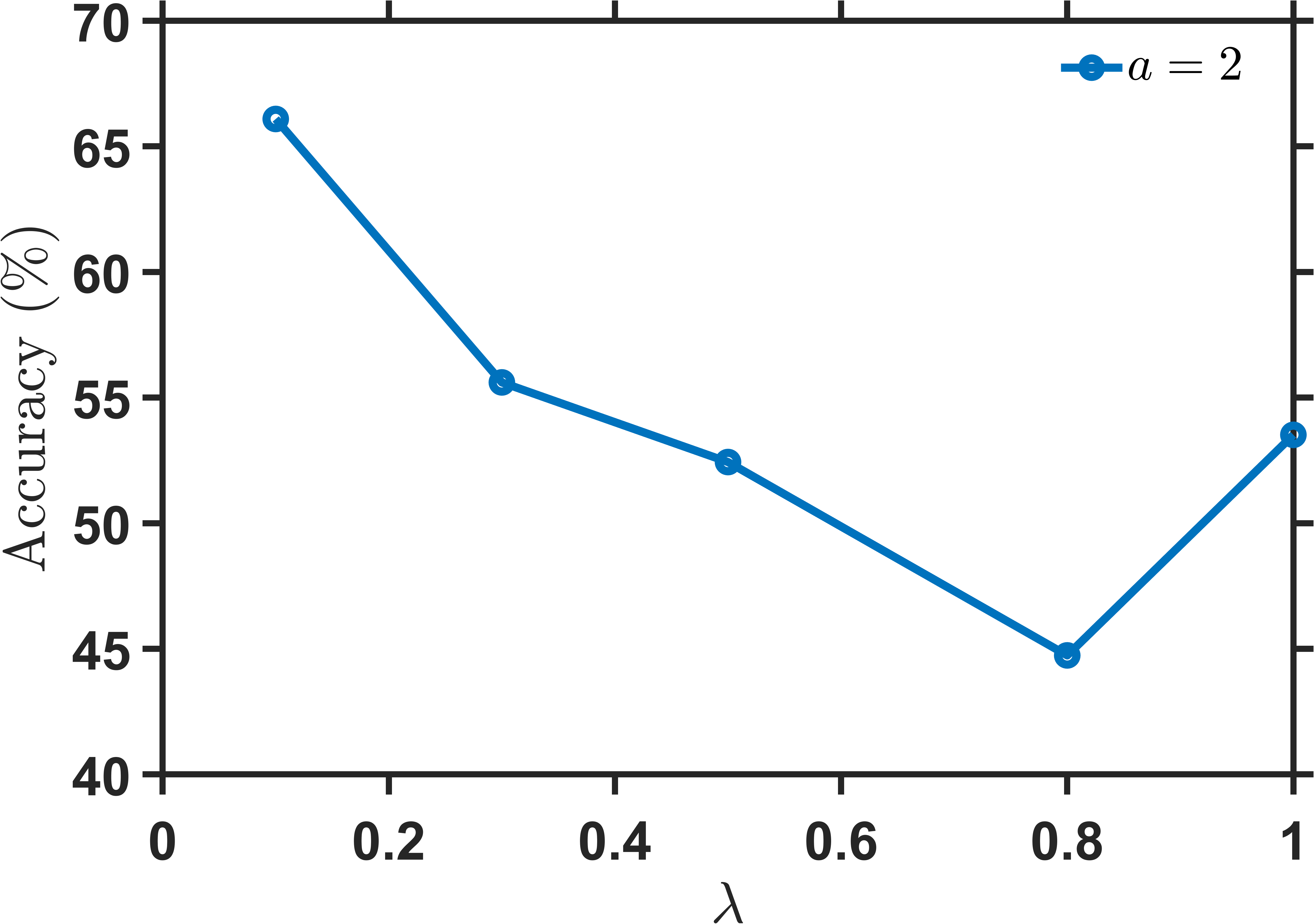}
        \label{fig:bc}
    }\hfill
    \subfloat[]{
        \includegraphics[width=0.30\textwidth]{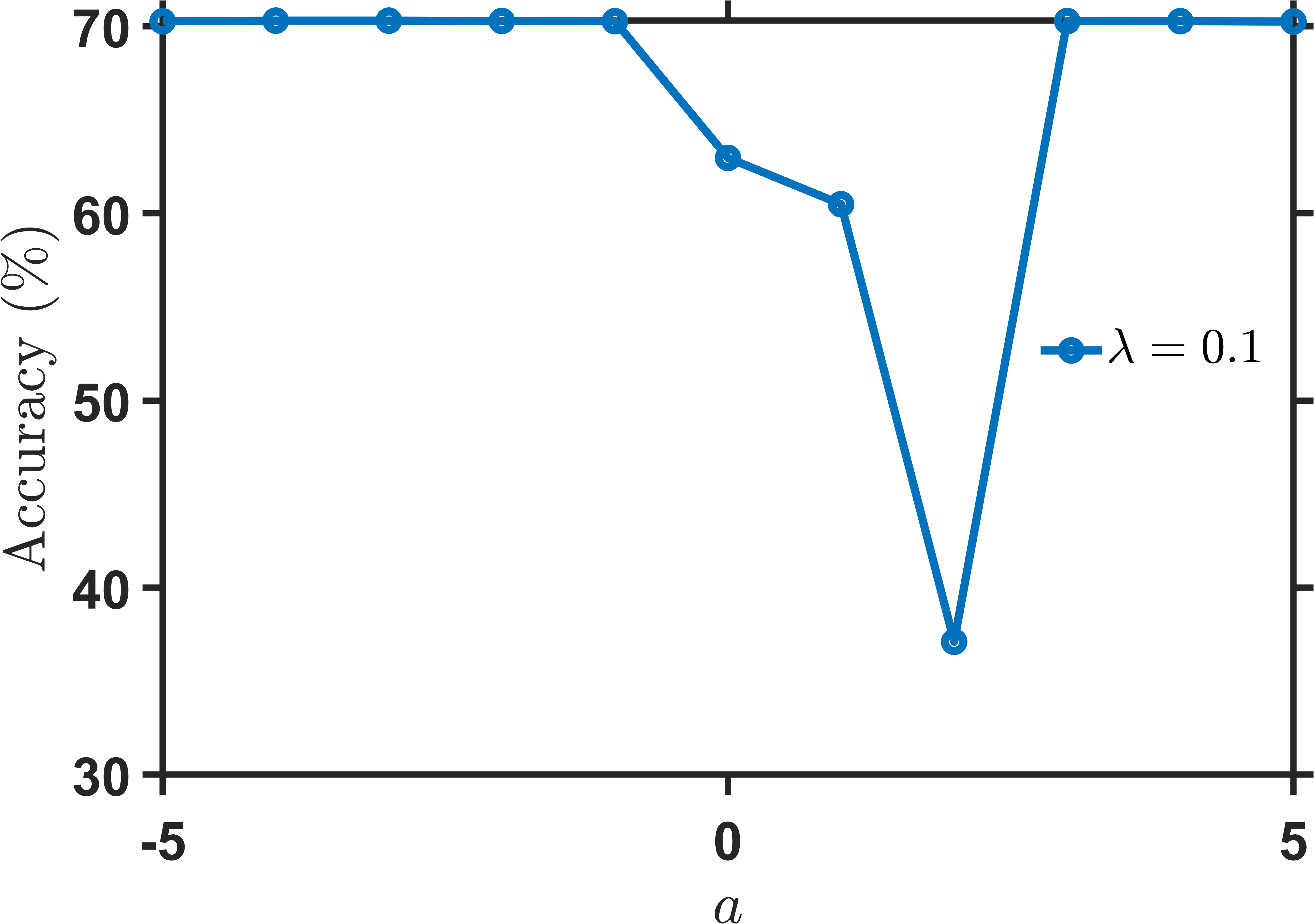}
        \label{fig:bcw}
    }\hfill
    \subfloat[]{
        \includegraphics[width=0.30\textwidth]{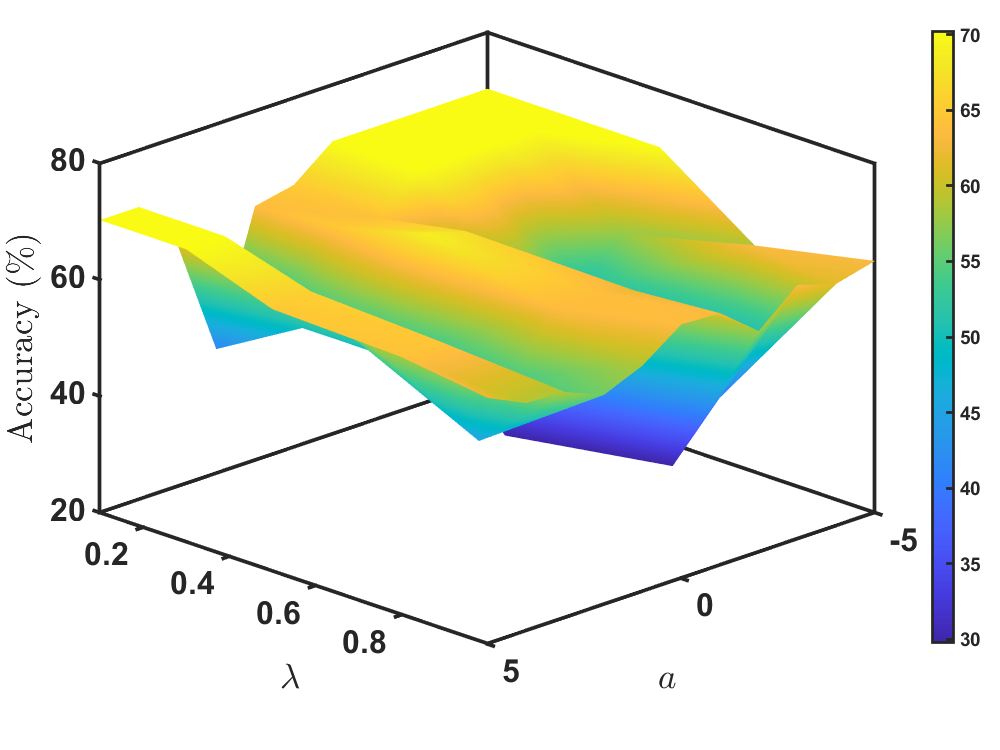}
        \label{fig:hv}
    }
\caption{Sensitivity analysis of the wave loss hyperparameters in the proposed Wave-BLS model on the breast\_cancer dataset. (a) Effect of the bounding parameter $\lambda$,
(b) effect of the shape parameter $a$,
and (c) their joint influence on classification accuracy.}
\end{figure*}
\section{Sensitivity Analysis of Wave Loss Hyperparameters}

In this subsection, we conduct a sensitivity analysis on the breast\_cancer dataset with respect to the wave loss hyperparameters. Specifically, we analyze the influence of the bounding parameter $\lambda$, the shape parameter $a$, and their joint interaction on classification accuracy, while fixing all remaining model hyperparameters at their optimal values. Fig.~\ref{fig:bc} shows the variation in classification accuracy on the breast\_cancer dataset as the bounding parameter $\lambda$ is varied while fixing $a$. Although the exact accuracy trend may differ across datasets, the results clearly indicate that the choice of $\lambda$ critically influences model performance. In particular, intermediate values of $\lambda$ lead to improved accuracy, whereas very small or large values result in performance degradation. This observation highlights the role of $\lambda$ in controlling the saturation behavior of the wave loss and regulating the influence of large errors, underscoring the importance of selecting an appropriate bounding parameter for stable and robust learning. Fig.~\ref{fig:bcw} depicts the sensitivity of accuracy with respect to the asymmetry parameter $a$ while keeping $\lambda$ fixed. For this dataset, Wave-BLS achieves consistently high accuracy over a broad range of both negative and positive values of $a$, indicating that the model is not overly sensitive to the choice of the asymmetry parameter. A sharp degradation in performance is observed only at a specific intermediate value of $a$, suggesting that certain configurations may induce unfavorable skewness in loss penalization. Nevertheless, the rapid recovery of accuracy for neighboring values highlights that Wave-BLS remains largely stable with respect to variations in $a$ when other hyperparameters are fixed. Fig.~\ref{fig:hv} presents the joint sensitivity surface of accuracy with respect to $a$ and $\lambda$. The surface reveals a broad region of relatively high accuracy, interspersed with localized valleys corresponding to unfavorable parameter combinations. This behavior indicates that, for this dataset, Wave-BLS does not require extremely precise tuning of $a$ and $\lambda$ to achieve competitive performance. At the same time, the presence of distinct low-accuracy regions highlights the importance of jointly selecting these parameters, as their interaction directly influences the balance between asymmetry and bounded error penalization.

Overall, this sensitivity analysis demonstrates that the wave loss hyperparameters play a meaningful role in shaping the performance of Wave-BLS. While the model exhibits a reasonable degree of stability across wide ranges of both $a$ and $\lambda$, suboptimal parameter choices can still lead to noticeable performance degradation. These observations suggest that a careful and informed selection of the wave loss parameters is essential to fully exploit its robustness properties and achieve consistently strong performance across datasets.


\end{document}